\documentclass[sigconf]{acmart}
 \pdfoutput=1
\usepackage{multirow}
\usepackage{makecell}
\usepackage{enumitem}
\usepackage{amsmath, bm}

\AtBeginDocument{%
  }

\setcopyright{acmcopyright}
\copyrightyear{2022}
\acmYear{2000}
\acmDOI{XXXXXXX.XXXXXXX}

\acmConference[Conference acronym 'XX]{Make sure to enter the correct
  conference title from your rights confirmation emai}{June 03--05,
  2018}{Woodstock, NY}
\acmPrice{15.00}
\acmISBN{978-1-4503-XXXX-X/18/06}

\begin{document}

\title{Exploring the Effectiveness of Video Perceptual Representation in Blind Video Quality Assessment}

\author{Liang Liao}
\affiliation{%
  \institution{S-lab, NTU, Singapore}
    \country{}
}
\email{liang.liao@ntu.edu.sg}

\author{Kangmin Xu}
\affiliation{%
  \institution{School of Computer Science, WHU, China}
    \country{}
}
\email{xukangmin@whu.edu.cn}

\author{Haoning Wu}
\affiliation{%
  \institution{S-lab, NTU, Singapore}
    \country{}
}
\email{haoning001@e.ntu.edu.sg}

\author{Chaofeng Chen}
\affiliation{%
  \institution{S-lab, NTU, Singapore}
    \country{}
}
\email{chaofeng.chen@ntu.edu.sg}

\author{Wenxiu Sun}
\author{Qiong Yan}
\affiliation{%
  \institution{Sensetime Research and Tetras AI}
    \country{}
}
\email{irene.wenxiu.sun@gmail.com}
\email{sophie.yanqiong@gmail.com}

\author{Weisi Lin}
\affiliation{%
  \institution{S-lab, NTU, Singapore}
    \country{}
}
\email{wslin@ntu.edu.sg}

\renewcommand{\shortauthors}{L. Liao et al.}

\begin{abstract}
With the rapid growth of in-the-wild videos taken by non-specialists, blind video quality assessment (VQA) has become a challenging and demanding problem. Although lots of efforts have been made to solve this problem, it remains unclear how the human visual system (HVS) relates to the temporal quality of videos. Meanwhile, recent work has found that the frames of natural video transformed into the perceptual domain of the HVS tend to form a straight trajectory of the representations. With the obtained insight that distortion impairs the perceived video quality and results in a curved trajectory of the perceptual representation, we propose a temporal perceptual quality index (TPQI) to measure the temporal distortion by describing the graphic morphology of the representation. Specifically, we first extract the video perceptual representations from the lateral geniculate nucleus (LGN) and primary visual area (V1) of the HVS, and then measure the straightness and compactness of their trajectories to quantify the degradation in naturalness and content continuity of video. Experiments show that the perceptual representation in the HVS is an effective way of predicting subjective temporal quality, and thus TPQI can, for the first time, achieve comparable performance to the spatial quality metric and be even more effective in assessing videos with large temporal variations. We further demonstrate that by combining with NIQE, a spatial quality metric, TPQI can achieve top performance over popular in-the-wild video datasets. More importantly, TPQI does not require any additional information beyond the video being evaluated and thus can be applied to any datasets without parameter tuning. Source code is available at \url{https://github.com/UoLMM/TPQI-VQA}.
\end{abstract}



\maketitle

\section{Introduction}
Recently, video streams have exploded on social media platforms, and most of them are captured by users in the wild with portable mobile devices~\cite{facebook2022,youtube2022,Tiktok2022}. Compared to videos from professionals, in-the-wild videos usually suffer from complicated distortion issues such as out of focus, over/under-exposure, and camera shake. Therefore, it is highly desirable to have an automatic quality assessment to eliminate low-quality videos or improve them during the acquisition and enhancement process. As these videos do not have pristine counterparts, a blind video quality assessment (VQA) is required. Although significant algorithms for blind images quality assessment (IQA) have been proposed~\cite{niqe2013,BRISQUE2012,HIGRADE17,zhang2021uncertainty,Ma2021active}, for videos, the temporal-domain quality is another integral aspect of blind VQA, since video perception is highly correlated with motion and temporal variations, leading VQA a more challenging problem.

\tabcolsep=0.5pt
\begin{figure*}[t]
	\centering
		\begin{tabular}{cccccc}
		    \multicolumn{2}{c}{\includegraphics[width=0.33\textwidth]{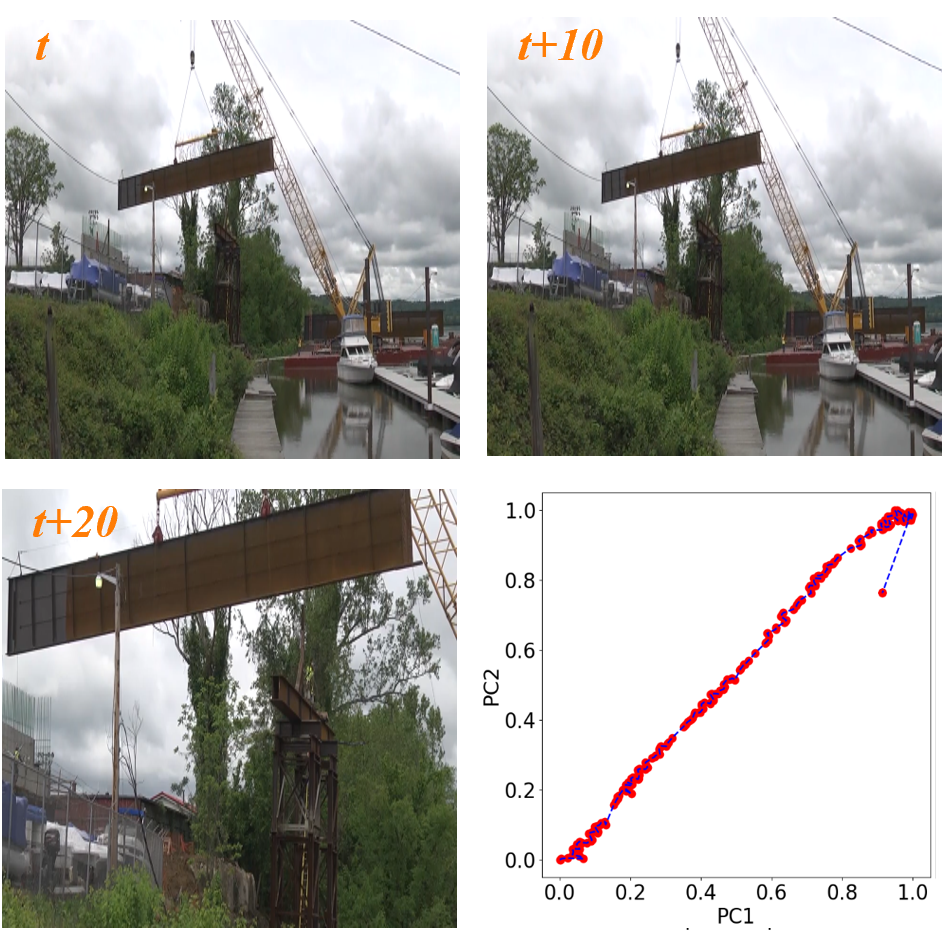}}~ &
		    \multicolumn{2}{c}{\includegraphics[width=0.33\textwidth]{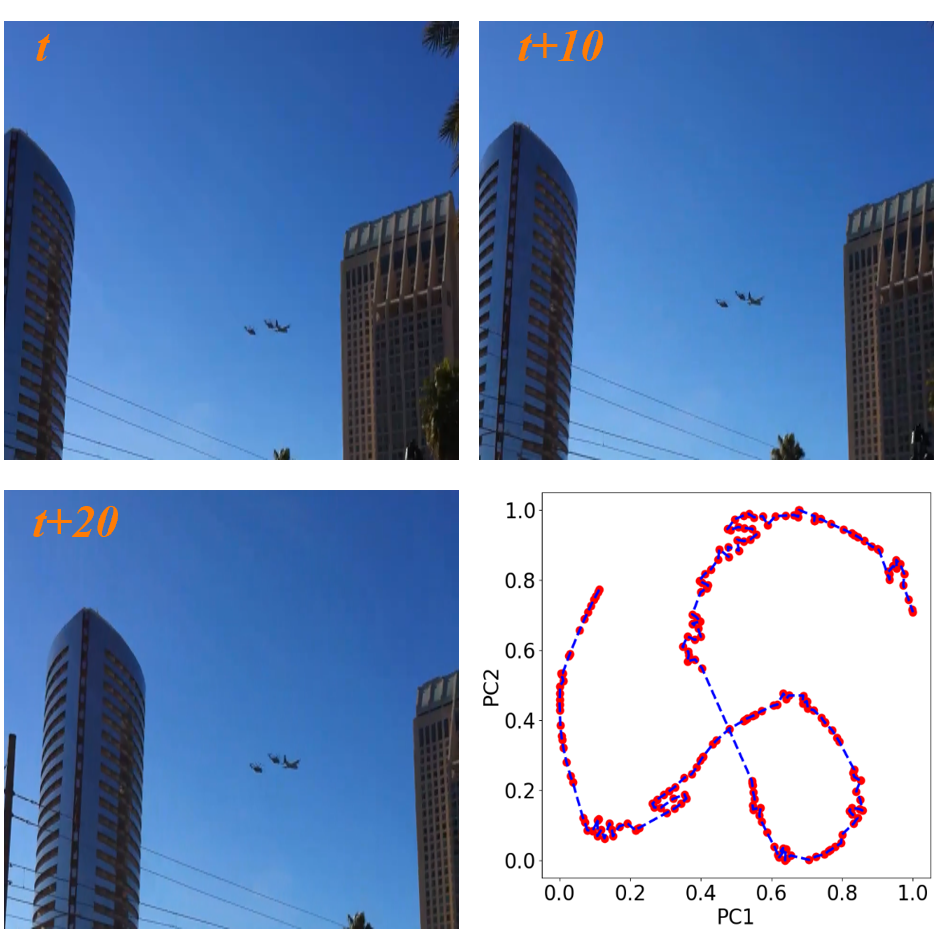}} ~&
		    \multicolumn{2}{c}{\includegraphics[width=0.33\textwidth]{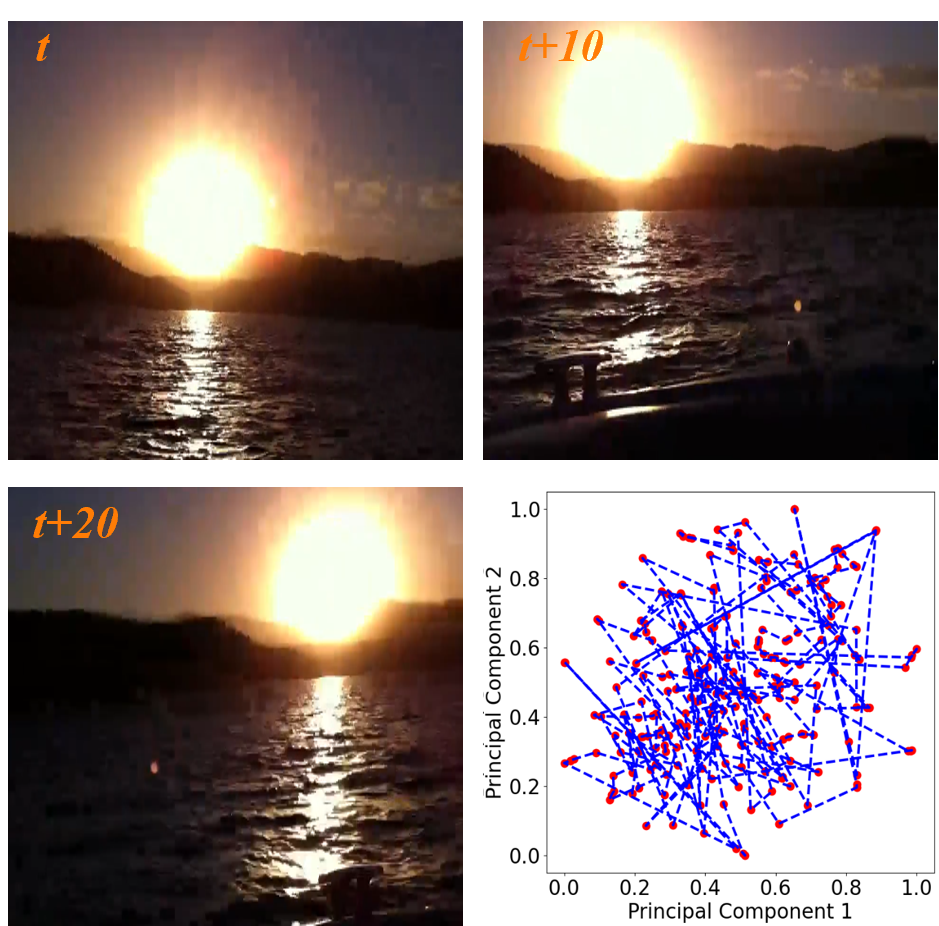}}\\
		    \multicolumn{2}{c}{(a) Video clips with MOS = 4.14} &
		    \multicolumn{2}{c}{(b) Video clips with MOS = 3.00} &
		    \multicolumn{2}{c}{(c) Video clips with MOS = 2.16}\\
		    \multicolumn{3}{c}{\includegraphics[width=0.495\textwidth]{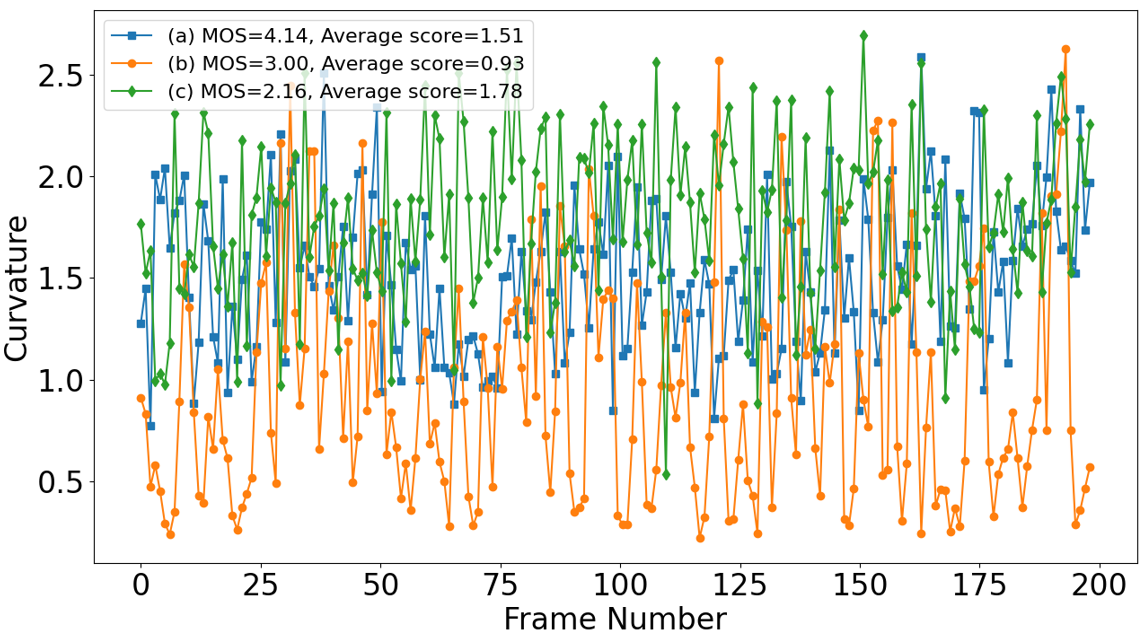}} &
		    \multicolumn{3}{c}{\includegraphics[width=0.495\textwidth]{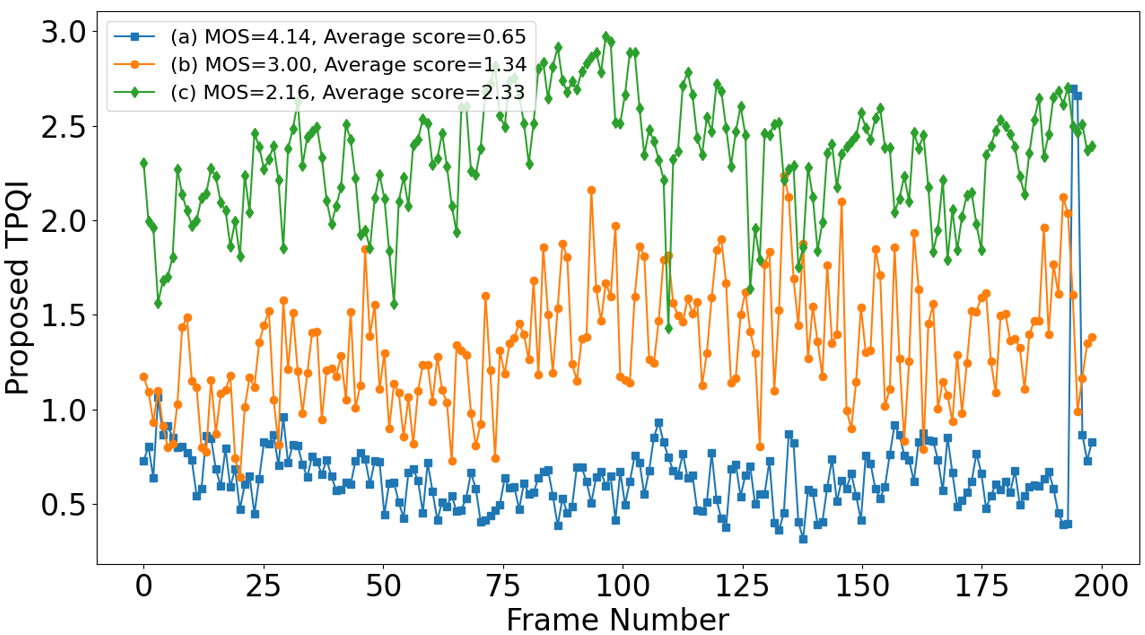}} \\
		    \multicolumn{3}{c}{(d) Quality scores calculated by Curvature~\cite{nature19} } &
		    \multicolumn{3}{c}{(e) Quality scores calculated by our VPT Descriptor} \\
	\end{tabular}
	\vspace{-2mm}
	\caption{\small Examples of three video clips from \textbf{KoNViD-1k} dataset. The chart next to the frames is visualized temporal trajectory (the two principal components) of all frames in the LGN domain. From (a) to (c), the temporal trajectory changes from nearly straight for high quality to curves for medium quality and to fragments for low quality. (d) and (e) show the distribution of quality scores measured by the curvature~\cite{nature19} and the proposed TPQI. In (d), the curves are mixed together and the average scores of the curvature do not have clearly relation with the MOS values; whilst in (e), the proposed TPQI scores are better separated for different qualities and inversely related to the MOS values.}
	\vspace{-2mm}
\label{motivation}
\end{figure*}


Up till now, some efforts have been made in modeling the quality in the temporal domain for VQA. One direct way is to compute frame-level quality scores and then express their relative importance over time by applying temporal pooling to these frame-level quality scores~\cite{icippooling20,VIDEVAL21,LSTMMM21}. However, the temporal pooling of spatial quality ignores the motion among frames. 
The most popular approach to temporal quality modeling is to deploy regular parametric bandpass models of natural scene statistics (NSS), such as 3D discrete cosine transform (3D-DCT)~\cite{3DDCT16} and 3D mean-subtracted contrast-normalized (3D-MSCN) coefficients~\cite{VIIDEO16,NSTSS20}, which characterize the perceived quality degradation by predicting the deviation of the distribution of frame-difference coefficients in the presence of distortion. 
However, in-the-wild videos contain authentic and commonly intermixed distortions, 
making NSS-based models designed for one specific distortion in each video unsuitable. 
Inspired by standout performance on a wide variety of computer vision tasks, deep learning-based VQA models~\cite{VSFA19,RIRNET20,GENERST21,STREPRE21,LSTMMM21,discovqa} have been proposed to extract content-aware and distortion-sensitive features to predict the quality of in-the-wild videos. However, most of them still integrate the frame-wise deep features by some pooling modules such as gated recurrent unit (GRU)~\cite{VSFA19} and long short-term memory (LSTM)~\cite{LSTMMM21}, making their performance constrained without effective extraction of motion information for video quality perception.
Although some work has been done on temporal quality modeling of video, research in this area is still in its infancy. The great challenge is that it is still unclear how humans perceive temporal distortion, especially for the in-the-wild videos.

Recent researches have reported the discovery on the straightness of the perceptual representation of natural videos~\cite{nature19, nature21}. 
It demonstrated that HVS transforms the incoming natural video signals into more regular representations in the perceptual domains, which are aligned along straighter trajectories in time. Examples of temporal trajectories of video clips, which track the perceptual representation of each frame in HVS along time, are visualized in Fig.~\ref{motivation}(a)-(c) (red point is the two principal components of the representation and dashed blue line is the temporal trajectory). The temporal trajectory is close to a straight line for the video with a high subjective quality score, but degenerates to haphazard curves for videos with low subjective quality scores. However, although the discovery has been successfully used to discriminate the natural and unnatural videos, \emph{i.e.} artificial and naturalistic sequences, the quality scores, predicted based on the curvature index induced in the temporal trajectories~\cite{nature19}, 
do not have clear relation with the human subjective scores for VQA (Fig.~\ref{motivation}(d)). We suspect that this may be due to only measuring the extent of straightness of the temporal trajectory is not enough for VQA. 




In this paper, we make the hypothesis that video distortions that harm the perceived quality of the videos will result in curved representations in HVS and propose a generalized and completely blind \textbf{T}emporal \textbf{P}erceptual \textbf{Q}uality \textbf{I}ndex (TPQI), through measuring the graphical morphology of the temporal representations of the videos in HVS (Fig.~\ref{motivation}(e)). Specifically, we first employ two computational models of HVS simulating the neural activity in the lateral geniculate nucleus (LGN) and primary visual area (V1) to transform the videos into their neural temporal trajectory representations. Then, we design a video perceptual trajectory (VPT) descriptor to quantify the temporal distortions. Considering that the distortions affect both the orientation and the fragmentation of the temporal trajectory, the VPT descriptor integrates two morphology elements into the measurement, \emph{i.e.} change of the direction and distance of the change. Experiments on various combinations of the elements are conducted to determine the final VPT descriptor.

We have evaluated the proposed TPQI on four popular in-the-wild video datasets and the results demonstrate that the proposed TPQI achieves almost the same performance as the well-established spatial quality index, NIQE~\cite{niqe2013}, indicating that the temporal quality index can be the same effective as the spatial quality index for VQA. Moreover, the representations of HVS in TPQI outperform the deep features from convolutional neural networks (CNNs), including Alexnet~\cite{Alexnet}, VGG~\cite{vgg19}, and Resnet~\cite{Resnet}, which are considered as candidate models for biological vision, showing that the VPT descriptor can well represent the neural perception of the temporal distortions. By integrating with NIQE, we can achieve state-of-the-art performance in the area of completely blind VQA, and even better than some of the opinion-aware blind VQA methods.
The main contributions of our paper can be summarized in threefold:
\begin{itemize}[leftmargin=0.75cm]
\item We exploit to characterize the complex temporal distortions of in-the-wild videos in the perceptual domain and demonstrate that the graphical morphology of the temporal trajectory representation can be treated as the indication of temporal distortion.
\item We propose a generalized and completely blind Temporal Perceptual Quality Index (TPQI) to measure the perceived temporal quality of video data by quantifying the loss of straightness and compactness of the temporal trajectory with a newly designed video perceptual trajectory descriptor.
\item We show for the first time that only the proposed TPQI can achieve VQA performance comparable to that of the spatial quality index and combining the spatial quality index and TPQI can achieve top performance over popular in-the-wild video datasets.
\end{itemize}

\section{Related Work}

\subsection{Video Quality Assessment}
Early VQA methods were specifically designed for synthetic distortions (\emph{e.g.}, Gaussian blur, compression, and transmission artifacts) based on statistical characteristics of the video, such as frame difference~\cite{BLIINDS14,NSTSS20}, gradient~\cite{GRADIENT19}, and optical flow~\cite{FLOW16,SALIENT19}. The most popular algorithms deploy perceptually relevant low-level features captured from natural bandpass statistical models. Typical approaches include Video-BLIINDS~\cite{BLIINDS14}, which uses a combination of temporal features from block-based motion estimation and DCT coefficients computed from frame differences, and spatial features from NIQE~\cite{niqe2013}, GM-LOG~\cite{GMLOG14}, which computes the joint statistics of gradient magnitude and Laplacian of Gaussian responses in the spatial domain, DESIQUE~\cite{DESIQUE13} in log-derivative and log-Gabor domains, and HIGRADE~\cite{HIGRADE17} in LAB color-transformed gradient domain. These methods estimate the deviations in the statistical distribution as a perceptual quality metric and have achieved good performance in quality assessment of synthetic distorted videos. However, their performance decreases significantly when applied to in-the-wild videos containing multiple unknown distortions.

Quality assessment of in-the-wild videos has attracted great attention given its potential broader practical utility. Attempting to capture the unknown and highly diverse distortions as possible as they can, recently proposed models used dozens of such perceptually relevant features and achieved state-of-the-art performance on existing in-the-wild datasets. For example, VIDEVAL~\cite{VIDEVAL21} is a bag of features-based blind VQA model on \textbf{KoNViD-1k} and \textbf{YouTube-UGC}, which uses a feature ensemble and selection procedure on top of existing efficient blind VQA models. RAPIQUE~\cite{RAPIQUE21} combines and exploits the advantages of both quality-aware scene statistics features and semantics-aware deep convolutional features, designing a general and efficient spatial and temporal bandpass statistical model for VQA. Instead of extracting handcraft features, deep VQA methods ~\cite{DNNVQA19,VSFA19,RIRNET20,RICHFEA21,Patch-VQ21,fastvqa} use CNNs to extract rich semantic features and run regression on the extracted features to predict video quality. For example, MLSP-FF~\cite{KONVID150K} extracts  frame-wise features with Inception-ResNetv2 model~\cite{Inception17} and some works~\cite{DNNVQA19,RICHFEA21,Patch-VQ21} introduce 3D-CNN instead of 2D-CNN to extract more efficient temporal features. Along with more powerful feature extraction, deep VQA methods attempt to achieve the temporal-memory effect by some temporal regression modules, such as recurrent neural network (RNN)~\cite{RIRNET20}, GRU~\cite{VSFA19} and LSTM~\cite{RICHFEA21,DNNVQA19}.

\tabcolsep=0.5pt
\begin{figure*}[htb]
	\centering
	\includegraphics[width=0.96\textwidth]{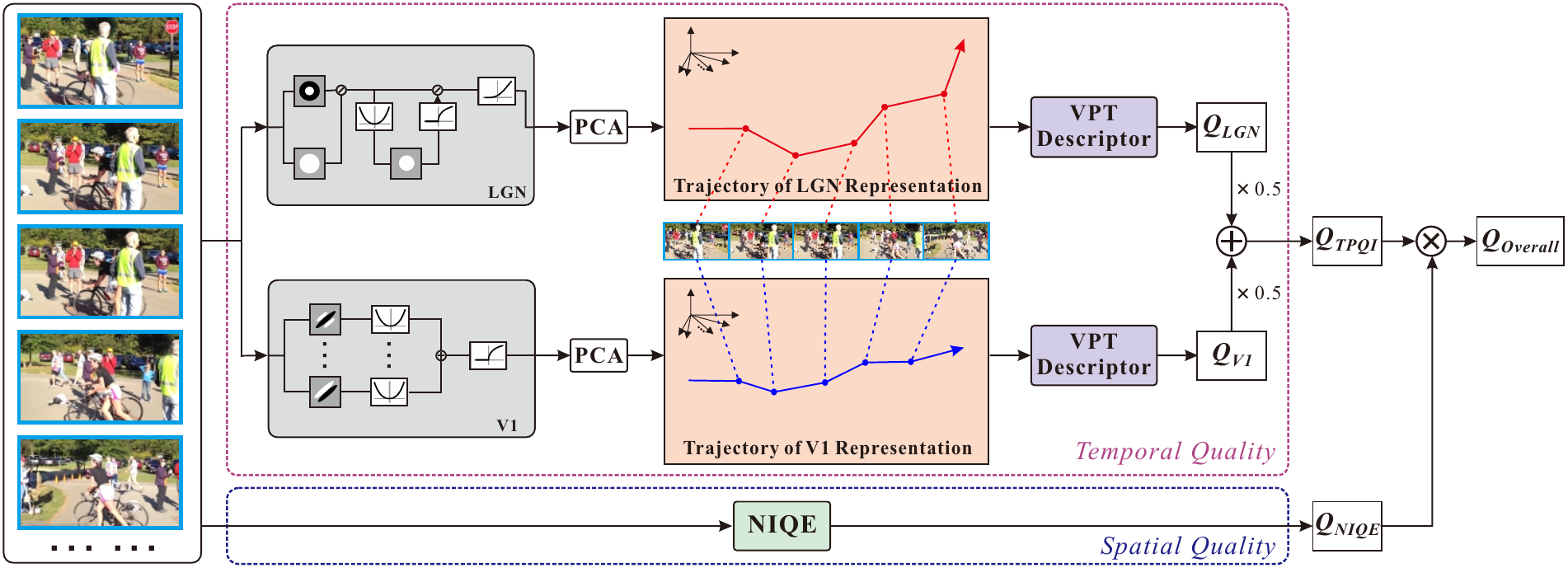} 
	\vspace{-3mm}
   \caption{\small Framework of the proposed completely blind VQA method. Each individual video frame is first transformed and obtained the temporal trajectory in perceptual domain, \textit{i.e.} LGN and V1 domains. A video perceptual trajectory descriptor is used to quantify temporal quality from both perceptual domains. The fusion of the temporal and spatial quality is used to predict the overall quality estimation.}
	\vspace{-3mm}
\label{fig:framework}
\end{figure*}

\subsection{Completely Blind Quality Assessment}
Most IQA/VQA methods require a large number of distorted images or videos with human subjective scores to learn the quality regression model, which leads to a massive workload in collecting these annotations. More importantly, it is difficult to collect enough training samples to cover the numerous distortion types, resulting in a weak generalization of opinion-aware quality assessment.

A few works have been carried out on completely blind IQA, \emph{i.e.} assessing quality without any additional information. Mittal \textit{et al.}~\cite{Mittalbiqa2012} first proposed a probabilistic latent semantic analysis of the statistical features of the pristine and distorted image patches and discovered latent quality factors that can infer a quality score from the test image patches. Later, 
they~\cite{niqe2013} proposed the Natural Image Quality Evaluator (NIQE), which infers the quality of a test image by measuring the distance between its Multivariate Gaussian (MVG) model learned from a set of local features from the image and the MVG model learned from the pristine natural images. Inspired by NIQE, Zhang \textit{et al.}~\cite{LINIEQ2015} replaced the local features with various NSS features and computed them from a collection of pristine image patches instead of the whole image. Liu \textit{et al.}~\cite{GUKE2020} proposed an unsupervised NR-IQA model based on the free energy principle to quantify the image quality in terms of structure, naturalness, and perceived quality changes during the degradation of test image.

As for the completely blind VQA, Mittal \textit{et al.}~\cite{VIIDEO16} proposed Video Intrinsic Integrity and Distortion Evaluation Oracle (VIIDEO), which was built on an NSS model of consecutive frame differences and measured departures from the statistical regularities in natural videos. Kancharla \textit{et al.}~\cite{STEM22} assumed that the increase in the straightness of perceptual domain representation was positively related to MOS values and performed a linear prediction in the LGN domain to model temporal distortions by the prediction errors.

\section{The Proposed Method}
We propose a video quality assessment algorithm based on the perceptual domain representation of the video extracted using bandpass models of the visual system, aiming to explore the mapping from the temporal trajectory of a video to the subjective perception of temporal quality. The framework of the proposed method is shown in Fig.~\ref{fig:framework}. We first transform each individual video frame into the perceptual domain, \textit{i.e.} LGN and V1 responses in this work, and obtain the temporal trajectory of the video by arranging the perceptual features along time. Then, we propose a video perceptual trajectory descriptor (VPT) to jointly measure the straightness and compactness of the trajectory, with the former modeling the distortions from naturalness of video and the latter modeling the content continuity between video frames. The temporal perceptual quality index (TPQI) is the average score from the VPT descriptors on both the LGN and V1 features. Finally, we combine the proposed TPQI with a spatial quality metric to predict the overall video quality.

\subsection{Perceptual Domain Representation of HVS}
We first extract the perceptual domain representations simulating the responses of LGN and V1, which are vital regions of the HVS for visual information processing. Specifically, LGN performs luminance and contrast gain control, while V1 is known to be tuned to different orientations, scales, and frequencies. As stated in~\cite{nature19}, the LGN representation likely straightens natural videos by providing robustness to local fluctuations in luminance and contrast, whereas the V1 representation provides straightening by its position- and phase-invariant properties. We use both representations to mimic the nonlinear functional properties of the early visual system.

\subsubsection{Extracting LGN representation}
The LGN model consists of center-surround filtering followed by local luminance and contrast gain control operations to simulate the primary nonlinear transformations performed by the retina and lateral geniculate nucleus~\cite{LGN2008functional}. In this work, we employ the LGN blocks as proposed by Laparra \textit{et al.}~\cite{lgnmodel16}. In this model, the linear components for luminance subtraction are implemented using difference of Gaussian (DoG) filters and a Laplacian pyramid. The non-linearity components perform contrast gain control to capture the local gain control property of the LGN neurons, which is achieved by performing a contrast normalization operation on the output of the linear bandpass filters. To relieve the tunning of hyper-parameters of the proposed model, we keep the settings of the LGN model as they are in~\cite{STEM22}.


\subsubsection{Extracting V1 representation}
The V1 model aims to transform the visual signal using a set of oriented filters whose responses are squared and combined over phases to capture the nonlinear behavior of complex cells in the primary visual cortex (V1)~\cite{V1985spatiotemporal}. In this work, we adopt Gabor filters as the simulation model of V1, motivated by the HVS hypothesis that Gabor filters have a good approximation of the response of V1~\cite{gabor1987}. Specifically, 
in the spatial domain, the 2-D Gabor filter is a Guassian kernel function modulated by a complex sinusoidal plane wave, defined as:
\begin{equation}
\begin{aligned}
    g_\theta(x,y)=\frac{f^2}{\pi\gamma\eta}exp&(-\frac{x^{\prime2}+\gamma^2y^{\prime2}}{2\sigma^2})exp(j2\pi fx^\prime+\phi),\\
    x^\prime=&x\cos{\theta}+y\sin{\theta},\\
    y^\prime=&-x\sin{\theta}+y\cos{\theta},\\
    \end{aligned}
    \vspace{1mm}
\end{equation}
where $f$ is the frequency of the sine wave, $\theta$ represents the orientation of the normal to the parallel stripes of the Gabor function, $\phi$ is phase offset, $\sigma$ is the standard deviation of the Gaussian envelope, and $\gamma$ is the spatial aspect ratio specifying the ellipticity of the support of the Gabor function. 
The Gabor filters are then used to convolve with each video frame, and the features from all Gabor filters are concatenated as the V1 representation of this frame.


\subsubsection{Dimensionality reduction of the perceptual domain representations}
To better understand the perceptual activities, many recent studies~\cite{pnas2015,bioRxiv2018,naturepca2021} have used dimensionality reduction techniques to transform the high-dimensional neural data into low-dimensional subspaces where the underlying manifolds are topologically simple. In this work, we apply principal component analysis (PCA) to reduce the representations of LGN and V1 to low-dimensional features of dimension $d$.

\subsection{Temporal Perceptual Quality Index}
\subsubsection{Motivation}

\begin{figure}[tb]
	\centering
	\small{
		\begin{tabular}{cccc}
			\includegraphics[width=0.48\columnwidth]{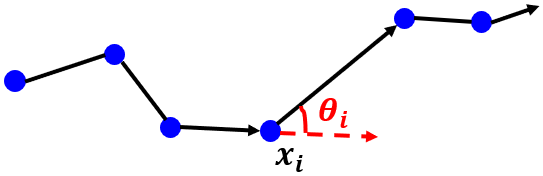} &
			\includegraphics[width=0.02\columnwidth]{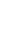} &
			\includegraphics[width=0.02\columnwidth]{./void.png} &
			\includegraphics[width=0.48\columnwidth]{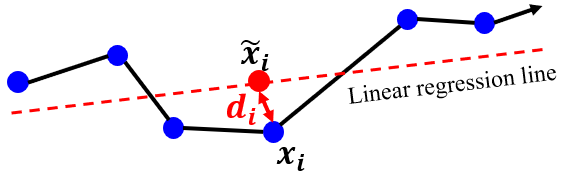} \\
			(a) Curvature index~\cite{nature19} &~&~& (b) Linear error			index~\cite{STEM22}\\
		\multicolumn{4}{c}{\includegraphics[width=0.5\columnwidth]{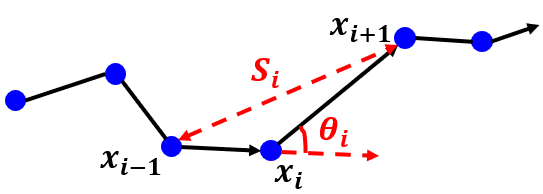}}\\
		\multicolumn{4}{c}{(c) The proposed VPT descriptor}\\
	\end{tabular}}
		\vspace{-4mm}

  \caption{\small Comparison of three kinds of trajectory descriptor for measuring temporal distortion. }
		\vspace{-4mm}
\label{fig:descriptor}
\end{figure}

Taking the temporal trajectories from LGN and V1 representations, we attempt to quantify their straightness and compactness losses to predict the temporal quality. A recent computational neuroscience model~\cite{nature19} defines the curvature of the trajectory as the average of the unsigned angles between difference vectors of successive frames to represent the straightness loss (Fig.~\ref{fig:descriptor}(a)). Although it is able to distinguish between natural and unnatural videos, it fails to map the curvature to the subjective quality scores (Fig.~\ref{motivation}(d)), probably because it does not quantify how far away a new frame deviates from the straight line, especially in the case of a large gap between two frames. Another alternative proposed in~\cite{STEM22} attempts to measure temporal distortion by calculating the distance between the predicted values from a linear model fit of the perceptual representations and the true representation (Fig.~\ref{fig:descriptor}(b)). It has taken the deviation distance from the trajectory into account, but ignores the variation between successive frames. 

In this paper, we propose to measure both the straightness and the compactness of the trajectory. The former measures the angular change between the difference vectors of successive frames, while the latter measures the degree of deviation between these frames (Fig.~\ref{fig:descriptor}(c)). In this way, if there is no directional change between two vectors, the magnitude of the vectors does not matter, but if there is, the distance will increase the penalty for the angular changes.




\subsubsection{Video perceptual trajectory descriptor}
In this work, we adopt three nearby frames as a temporal trajectory unit as in~\cite{nature19} and integrate two types of morphology change elements, including change of the direction and distance of the change to measure temporal distortion degree and thereby predict the temporal quality.

\textbf{Direction change:} The curvature is used to measure the direction of the change, which is defined as the angle between the nearby difference vectors. Specifically, let $\bm{x_{i-1}}$, $\bm{x_{i}}$ and $\bm{x_{i+1}}$ be the perceptual representations of three consecutive frames, which are located in a $d$-dimensional space after dimensionality reduction, the curvature can be calculated as:
\begin{equation}
    \bm{\overrightarrow{x_{i-1} x_{i}}} = \bm{x_{i}} - \bm{x_{i-1}},
\end{equation}
\begin{equation}
    \bm{\overrightarrow{x_{i} x_{i+1}}} = \bm{x_{i+1}} - \bm{x_{i}},
\end{equation}
\begin{equation}
    \theta_i = arccos(\frac{\bm{\overrightarrow{x_{i-1} x_{i}}}}{||\bm{\overrightarrow{x_{i-1} x_{i}}}||}\cdot\frac{\bm{\overrightarrow{x_{i} x_{i+1}}}}{||\bm{\overrightarrow{x_{i} x_{i+1}}}||}),
\end{equation}
where $\bm{\overrightarrow{x_{i-1} x_{i}}}$ and $\bm{\overrightarrow{x_{i} x_{i+1}}}$ denote the two difference vectors of the three consecutive frames. $\cdot$ denotes to vector dot product and $\theta_i$ is the curvature in radians.

\textbf{Distance change:} We use the magnitude of sum of the two difference vectors to measure the distance of the change. Specifically, the distance can be calculated as:
\begin{equation}
    S_i = ||\bm{\overrightarrow{x_{i-1} x_{i}}}+\bm{\overrightarrow{x_{i} x_{i+1}}}|| = ||\bm{x_{i+1}}- \bm{x_{i-1}}||.
\end{equation}

    \textbf{VPT descriptor:} Assuming both the extent of the change direction and distance are negatively correlated to the temporal quality, \emph{i.e.} the larger the change in direction or the longer the change in distance, the greater temporal distortion and thus the lower temporal quality, we define a generic representation of these two morphology change elements to measure the temporal distortion at each time instant $i$, given by:
\begin{equation}
    Q_i = f(\theta_i, S_i),
\end{equation}
where $f(\cdot)$ denotes a pairwise function of $\theta_i$ and $S_i$. In this work, we use $f(\theta_i, S_i)=\theta_i\times\sqrt{S_i}$ after experimenting with various choices.

\subsubsection{Calculating the score of TPQI}
To estimate the video-level temporal quality, the temporal averages of $Q_i$ in the LGN and V1 domains are calculated and followed by logarithmic compression as in~\cite{STEM22} separately, and then the temporal quality score of the whole video is obtained from the average of the two domains.
\begin{equation}
    Q_{TPQI}=\frac{\log({\frac{1}{N-2}\sum_{i=2}^{N-1}Q_i^{LGN}}) + \log({\frac{1}{N-2}\sum_{i=2}^{N-1}Q_i^{V1}}))}{2},
\end{equation}
where $N$ is the total number of video frames, and $Q_{TPQI}$ is calculated from frame 2 to frame $N-1$.


\tabcolsep=0.5pt

\subsection{Natural Video Quality Evaluator}
\subsubsection{Spatial quality estimation}
It is commonly acknowledged that frame-level spatial quality plays a very important role in estimating the overall video quality in the VQA problem. In our work, we employ the well-established blind NR-IQA algorithm, NIQE~\cite{niqe2013}, for spatial quality estimation. The overall spatial quality of a video is the average of the frame-level spatial quality scores, given by:
\vspace{-1mm}
\begin{equation}
    Q_{NIQE} = \frac{1}{N}\sum_{i=1}^{N}q_i,
\vspace{-1mm}
\end{equation}
where $q_i$ denotes to the spatial quality score of $i$-th frame and $N$ is the total frame number of the video.

\subsubsection{Overall video quality estimation}
The overall video quality estimate is the fusion of spatial and temporal quality given by:
\begin{equation}
    Q_{overall} = fusion(Q_{NIQE}, Q_{TPQI}),
\end{equation}
where $fusion(\cdot)$ denotes the fusion strategy. There are two fusion strategies of defining the overall quality index for a video, \textit{i.e.} average or product of spatial and temporal quality, which both cause the index to respond to percentage changes in either spatial or temporal indices. 
We study the effectiveness of the two fusion strategies in the experimental section.

\begin{table*}[htb]
    \centering
        \setlength\tabcolsep{4pt}
    \caption{\small Performance comparison on the four VQA datasets. TPQI (LGN), TPQI (V1) and TPQI (LNG+V1) are the three TPQI variants adopting temporal representations from different perceptual domains. Overall (Sum) and Overall (Product) denote two different strategies of fusing spatial and temporal quality scores.}
    \vspace{-3mm}
    \linespread{0.95}
    \resizebox{\linewidth}{!}{
    \begin{tabular}{c|c|ccc|ccc|ccc|ccc}
    \toprule
   \multirow{2}{*}{Category}  &\multirow{2}{*}{Method}  & \multicolumn{3}{c|}{\textbf{KoNViD-1k}}& \multicolumn{3}{c|}{\textbf{LIVE\_VQC}}& \multicolumn{3}{c|}{\textbf{CVD2014}}& \multicolumn{3}{c}{\textbf{YouTube-UGC}}\cr\cline{3-14}
   ~&~ & SRCC$\uparrow$ &PLCC$\uparrow$&RMSE$\downarrow$ & SRCC$\uparrow$ &PLCC$\uparrow$&RMSE$\downarrow$ & SRCC$\uparrow$ &PLCC$\uparrow$&RMSE$\downarrow$ & SRCC$\uparrow$ &PLCC$\uparrow$&RMSE$\downarrow$\\ \midrule
 \multirow{4}{*}{\makecell[c]{Opinion-aware \\ blind VQA}}  &V-BLIINDS~\cite{BLIINDS14} &0.710&	0.704&	0.460&	0.694&	0.718&	11.765&	0.700&	0.710&	15.222	&0.559&	0.555	&0.536 \\
 ~&TL-VQM~\cite{TLVQM19} &0.780&	0.770&	0.406& 0.799&	0.803&	10.145&	0.830&	0.850&	11.330&	0.669&	0.659&	0.485 \\
 ~&NSTSS~\cite{NSTSS20} &0.625&	0.639&	-&	-&	-&	-&	0.615&	0.653&	-&	-&	-&	- \\
~& VIDEVAL~\cite{VIDEVAL21} & 0.783&	0.780&	0.403&	0.752&	0.751&	11.100&	-&	-&	-&	0.779&	0.773&	0.405 \\ \midrule
 \multirow{3}{*}{\makecell[c]{Completely \\ blind VQA}}  & NIQE~\cite{niqe2013}&0.543&	0.548&	0.536&	0.598&	0.622&	13.356&	0.492&	0.612&	16.950&	0.266&	0.290&	0.640  \\
~& VIIDEO~\cite{VIIDEO16}&-0.015&	0.013&	0.639&	0.029&	0.137&	16.882&	0.145&	0.119&	23.644&	0.160&	0.146&	0.637  \\ 
 ~& STEM~\cite{STEM22} &0.629&	0.629&	0.497&	0.656&	0.670&	12.649&	0.532&	0.619&	16.813&	\textbf{0.284}&	\textbf{0.318}	&0.623 \\ \midrule
 \multirow{5}{*}{\makecell[c]{Completely \\ blind VQA \\ (Ours)}}  & TPQI (LGN) &0.453&0.439&0.576&0.519&0.516&14.071&0.229&0.356&20.243&0.047&0.064&0.635  \\
 ~& TPQI (V1) &0.531&0.527&0.545&0.596&0.618&13.276&0.408&0.469&19.057&0.269&0.311&0.617 \\
 ~& TPQI (LGN+V1) &0.556&0.549&0.541&0.636&0.645&12.907&0.413&0.464&19.107&0.111&0.218&0.626  \\
 ~& Overall (Sum) &0.660&0.659&0.482&0.708&0.721&11.714&\textbf{0.553}&\textbf{0.637}&\textbf{16.693}&0.268&0.297&\textbf{0.613}  \\
 ~& Overall (Product) &\textbf{0.693}&\textbf{0.693}&\textbf{0.462}&\textbf{0.718}&\textbf{0.730}&\textbf{11.550}&0.524&0.597&17.368&0.230&0.288&0.615 \\
    \bottomrule
    \end{tabular}
    }
    \vspace{-2mm}
 \label{sota}
\end{table*}

\section{Experimental Results}

\subsection{Experimental Settings}
\subsubsection{Datasets} We evaluate the effectiveness of the proposed method on four popular in-the-wild VQA datasets, including: \textbf{KoNViD-1k}, \textbf{LIVE-VQC}, \textbf{CVD2014}, and \textbf{YouTube-UGC}.

\textbf{KoNViD-1k~\cite{konvid17}}: The dataset contains 1,200 videos and the resolution of all videos is $960\times540$. The videos are 8 seconds in duration with frame rates of 24/25/30 frames per second (fps).


\textbf{LIVE-VQC~\cite{livevqc}}: The dataset contains 585 videos with more temporal variations than the \textbf{KoNViD-1k} dataset. The resolution of the videos ranges from 240P to 1080P, and the videos are 10 seconds in duration with frame rates ranging from 19 to 30 fps.

\textbf{CVD2014~\cite{CVD2014}}: The dataset contains 234 videos 
and the resolutions of the videos are 480P and 720P. The duration and frame rates range from 10 to 25 seconds and 11 to 31 fps, respectively.

\textbf{YouTube-UGC~\cite{youtubeugc}}: The dataset has 1,131 videos
with authentic distortions of 15 categories. The resolution of the videos varies from 360P to 4K, and all videos are 20 seconds in duration.


\subsubsection{Baseline Methods} For comparison, we select three completely blind methods, \emph{i.e.} NIQE~\cite{niqe2013}, VIIDEO~\cite{VIIDEO16}, and STEM~\cite{STEM22}. We also compare our proposed method with four representative opinion-aware blind VQA algorithms, including V-BLIINDS~\cite{BLIINDS14}, TLVQM~\cite{TLVQM19}, NSTSS~\cite{NSTSS20}, and VIDEVAL\cite{VIDEVAL21}. Unlike our proposed method that does not require any annotation, these opinion-aware algorithms need a training procedure that regresses various features extracted to the annotated MOS values. The numerical results of these baselines are presented from the literature~\cite{VIDEVAL21,NSTSS20,STEM22}.

\subsubsection{Evaluation Metrics} We report three widely used metrics to evaluate the VQA performance, including Spearman’s rank correlation coefficient (SRCC), Pearson’s linear correlation coefficient (PLCC), and root mean square error (RMSE). SRCC and PLCC measure the correlation between predicted quality scores and labeled MOS values, and RMSE indicates the relative error. A better VQA method would result in higher SRCC and PLCC, but lower RMSE.

Considering the inconsistency of the scale between the predicted quality scores and the subjective scores, we perform the nonlinear mapping with a 4-parameter logistic function as suggested by VQEG~\cite{VQEG2000}. The function is formulated as follows.
\begin{equation}
    Q_{fit} = \beta_2+\frac{\beta_1-\beta_2}{1+\exp{(\frac{-(Q_{pre}-\beta_3)}{|\beta_4|})}}
\end{equation}
where $Q_{pre}$ and $Q_{fit}$ denote the predicted score and mapped score, respectively. $\beta_1$, $\beta_2$, $\beta_3$ and $\beta_4$ are the four fitting parameters of the logistic function. 

\subsubsection{Implementation details} The PCA dimension is set to 10 and a spatial resolution of $480\times270$ for all videos to extract the V1 representations, which were both analyzed in the experiments. We employ 48 Gabor filters with 6 scales and 8 orientations and the size of the Gabor filters was set to $39\times39$.The raw spatial resolution was used for calculating NIQE.

\begin{table}[tb]
    \centering
        \setlength\tabcolsep{4pt}
    \caption{\small Numerical comparison on the performances of trajectory descriptors. Linear denotes the linear prediction error in~\cite{STEM22}. The features used for comparison are from both LGN and V1.}
    \linespread{0.95}
    \vspace{-2mm}
    \resizebox{\columnwidth}{!}{
    \begin{tabular}{clccccccc}
    \toprule
        ~& \multirow{2}{*}{Descriptor}  & \multicolumn{2}{c}{\textbf{KoNViD-1k}}& \multicolumn{2}{c}{\textbf{LIVE\_VQC}}& \multicolumn{2}{c}{\textbf{CVD2014}}\cr \cline{3-8}
        ~&  ~& SRCC$\uparrow$ & PLCC$\uparrow$& SRCC$\uparrow$ & PLCC$\uparrow$& SRCC$\uparrow$ & PLCC$\uparrow$ \\ \midrule
    \multirow{4}{*}{\rotatebox{90}{Temporal}}&Linear~\cite{STEM22} &0.504 &0.497 &0.567 &0.611&0.238 &0.410\\
    ~&Curvature &0.425 &0.423 &0.539 &0.548&0.380 &0.430\\
     ~&Distance &0.411 &0.403 &0.460 &0.505&0.310 &0.315  \\
    ~& Ours&  \textbf{0.556} &\textbf{0.549} &\textbf{0.636} &\textbf{0.645}&\textbf{0.413} &\textbf{0.464}   \\\midrule
    \multirow{4}{*}{\rotatebox{90}{Overall}}&Linear~\cite{STEM22} &0.642 &0.644 &0.662 &0.686&0.394 &0.450\\
    ~&Curvature &0.401 &0.429 &0.660 &0.677&0.479 &0.515  \\
     ~&Distance &0.632 &0.641 &0.666 &0.684&0.354 &0.359  \\
     ~&Ours&  \textbf{0.693} &\textbf{0.693} &\textbf{0.718} &\textbf{0.730}&\textbf{0.524} &\textbf{0.597} \\
    \bottomrule
    \end{tabular}
    }
    \vspace{-3mm}
 \label{tab:des}
\end{table}

\subsection{Performance Evaluation}
We first compare the performance with the baselines on four datasets. The results are shown in Table~\ref{sota} and analyzed in detail as follows. 

\textbf{Overall performance.} The proposed TPQI delivers competitive performance over all completely blind baselines, and the overall performance of the combined TPQI and spatial metric NIQE achieves the best performance over three datasets. The overall performance is even better than some opinion-aware VQA baselines, \emph{e.g.}, better than V-BLIINDS on \textbf{LIVE\_VQC} dataset and better than NSTSS on \textbf{KoNViD-1k} and \textbf{CVD2014} datasets. The results show that the proposed TPQI does not require any dataset-specific information, and can be generalized to any video with natural settings.

\begin{table*}[tb]
    \centering
        \setlength\tabcolsep{5pt}
    \caption{\small Numerical comparison on the performances of different choices of change distance in the trajectory unit. ($\bm{x_{i+1}}\to\bm{\protect\overrightarrow{x_{i-1}x_{i}}}$ means the distance from the representation $\bm{x_{i+1}}$ to vector $\bm{\protect\overrightarrow{x_{i-1}x_{i}}}$).}
    \vspace{-3mm}
    \linespread{0.95}
    \resizebox{0.85\linewidth}{!}{
    \begin{tabular}{cc|cc|cc|cc|cc|cc}
    \toprule
    \multirow{2}{*}{Dataset} & \multirow{2}{*}{Domain}  & \multicolumn{2}{c|}{$||\bm{\overrightarrow{x_{i-1} x_{i}}}||$}& \multicolumn{2}{c|}{$||\bm{\overrightarrow{x_{i} x_{i+1}}}||$}& \multicolumn{2}{c|}{$||\bm{\overrightarrow{x_{i-1} x_{i}}}||+||\bm{\overrightarrow{x_{i} x_{i+1}}}||$}& \multicolumn{2}{c|}{$||\bm{\overrightarrow{x_{i-1} x_{i}}}+\bm{\overrightarrow{x_{i} x_{i+1}}}||$}& \multicolumn{2}{c}{\textbf{$\bm{x_{i+1}}\to\bm{\overrightarrow{x_{i-1} x_{i}}}$}}\cr\cline{3-12}
    ~&~ & SRCC$\uparrow$ &PLCC$\uparrow$ & SRCC$\uparrow$ &PLCC$\uparrow$ & SRCC$\uparrow$ &PLCC$\uparrow$ & SRCC$\uparrow$ &PLCC$\uparrow$& SRCC$\uparrow$ &PLCC$\uparrow$\\ \midrule
  \multirow{2}{*}{\textbf{KoNViD-1k}}& Temporal & 0.544&0.535&0.546&0.537&0.550&0.541&\textbf{0.556}& \textbf{0.549}&0.554&0.545\\
  ~& Overall &0.678&0.680 &0.680&0.681&0.690&0.690&\textbf{0.693}&\textbf{0.693}&0.673&0.677 \\\midrule
  \multirow{2}{*}{\textbf{LIVE-VQC}}& Temporal  &0.632&0.638& 0.632&0.639&0.635&0.641&\textbf{0.638}&\textbf{0.647}&0.636&0.645 \\
  ~& Overall &0.710&0.722 &0.711&0.723&0.717&\textbf{0.730}&\textbf{0.718}&\textbf{0.730}&0.697&0.714  \\ \midrule
  \multirow{2}{*}{\textbf{CVD2014}}&  Temporal &0.388&0.442 &0.386&0.441&0.396&0.452&0.413&0.464&\textbf{0.421}&\textbf{0.470}  \\
  ~&  Overall &0.522&0.584& 0.522&0.584&\textbf{0.544}&\textbf{0.609}&0.524&0.597&0.533&0.592  \\\midrule
  \multirow{2}{*}{\textbf{YouTube-UGC}}&  Temporal  &0.114&0.216& 0.115&0.217&0.115&0.218&0.111&\textbf{0.218}&\textbf{0.117}&0.216 \\
  ~& Overall &0.211& 0.291& 0.212&0.292&0.236 &\textbf{0.298}& 0.230& 0.288&0.211&0.265  \\
    \bottomrule
    \end{tabular}
    }
    \vspace{-3.5mm}
 \label{tab:distance}
\end{table*}

\begin{table}[tb]
    \centering
        \setlength\tabcolsep{5pt}
    \caption{\small  Numerical comparison of different video resolutions for V1 feature on \textbf{KoNViD-1k}. The time unit is second/frame on CPU.}
    \vspace{-3mm}
    \linespread{0.95}
        \resizebox{\columnwidth}{!}{
    \begin{tabular}{c|ccc|ccc}
    \toprule
    \multirow{2}{*}{\makecell[c]{Video Resolution\\(Downsam. rate)}} &\multicolumn{3}{c|}{TPQI (V1)}& \multicolumn{3}{c}{Overall}\cr \cline{2-4}\cline{5-7}
    ~&SRCC$\uparrow$ &PLCC$\uparrow$&Time & SRCC$\uparrow$ & PLCC$\uparrow$&Time\\ \midrule
    $960\times540$ (1)&0.522&0.518&0.928&0.695&0.694&1.234\\
    $480\times270$ (1/2)&0.531&0.527&0.254&0.693&0.693&0.334\\
    $240\times135$ (1/4)&0.521&0.520&0.067&0.680&0.682&0.095\\
    $120\times67$ (1/8)&0.501&0.501&0.033&0.662&0.665&0.044\\
    \bottomrule
    \end{tabular}
    }
    \vspace{-5mm}
 \label{resolution}
\end{table}

Notice that the proposed method does not reach the best performance on \textbf{YouTube-UGC} dataset, and the performance degradation is in accordance with the results of NIQE, which is constructed based on the statistical regularities of natural images. The reason may be that this dataset contains many unnatural video categories, such as \textit{Animation}, \textit{Gaming}, and \textit{Lyric Video}, which are subjected to artificial processing and do not conform to the straightness hypothesis in the perceptual domain.

\textbf{Employing features from different perceptual domains.} The comparison among TPQI (LGN), TPQI (V1) and TPQI (LGN+V1) studies the effectiveness of the perceptual representations from different domains
on measuring the temporal quality. The performance increases from the LGN domain to the V1 domain with the depth of the visual system, and the linear combination of LGN and V1 features can further boost the performance. The results show that both features play important roles in predicting subjective scores, and these features can also compensate for each other. They also show that our proposed TPQI, which measures only the temporal quality, can achieve better VQA performance than NIQE, especially on \textbf{LIVE\_VQC}, which includes large camera motions.

\textbf{Fusion of TPQI with spatial quality metric.} We conduct comparative experiments on the fusion strategies, \textit{i.e.} summation or multiplication, of the scores from the proposed temporal index (TPQI) and the spatial quality metric (NIQE). It can be observed that the product of the spatial and temporal scores leads to higher accuracy for \textbf{KoNViD-1k} and \textbf{LIVE\_VQC}, which is probably contributed by the relative insensitivity of our indices to the range of values occupied by the spatial and temporal indices. Thus we adopt the product strategy in the ablation study to obtain the overall quality predictions.


\tabcolsep=0.5pt
\begin{figure}[t]
	\centering
	\small{ 
		\begin{tabular}{c}
		    \includegraphics[width=0.9\columnwidth]{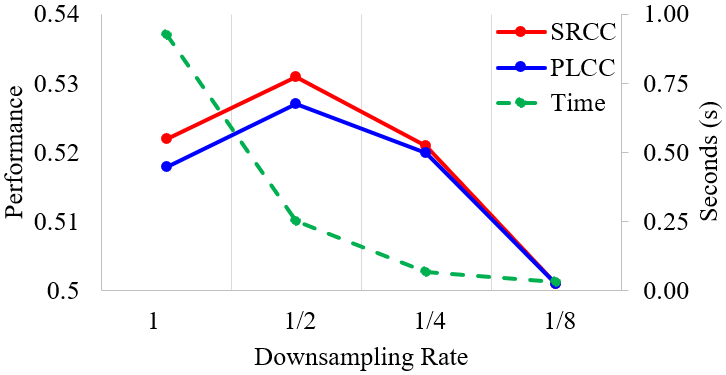} \\
		    (a) Comparison on TPQI (V1)\\
		    \includegraphics[width=0.9\columnwidth]{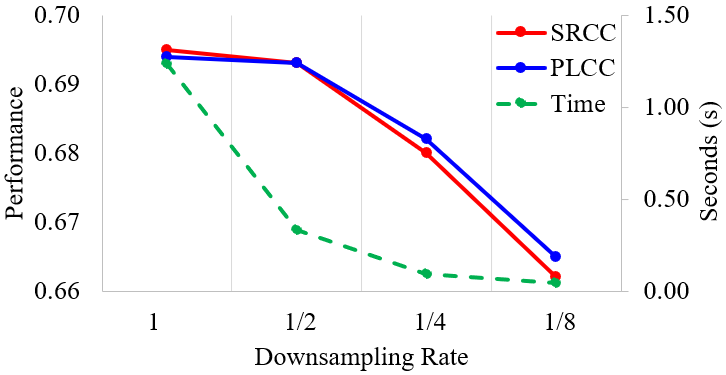} \\
		    (b) Comparison on overall quality\\
	\end{tabular} }
	    \vspace{-4mm}

\caption{\small Performance comparison of different video resolutions for V1 feature extraction on \textbf{KoNViD-1k}.}
    \vspace{-5mm}
\label{fig:resolution}
\end{figure}

\begin{table*}[tb]
    \centering
        \setlength\tabcolsep{5pt}
    \caption{\small Numerical comparison on the performances of different dimensions $d$ of the representation in V1 domain.}
    \vspace{-3mm}
    \linespread{0.92}
    \resizebox{0.85\linewidth}{!}{
    \begin{tabular}{cc|cc|cc|cc|cc|cc}
    \toprule
    \multirow{2}{*}{Dataset} & \multirow{2}{*}{Domain}  & \multicolumn{2}{c|}{\textbf{d=5}}& \multicolumn{2}{c|}{\textbf{d=10}}& \multicolumn{2}{c|}{\textbf{d=30}}& \multicolumn{2}{c|}{\textbf{d=50}}& \multicolumn{2}{c}{\textbf{d=80}}\cr\cline{3-12}
    ~&~ & SRCC$\uparrow$ &PLCC$\uparrow$ & SRCC$\uparrow$ &PLCC$\uparrow$ & SRCC$\uparrow$ &PLCC$\uparrow$ & SRCC$\uparrow$ &PLCC$\uparrow$& SRCC$\uparrow$ &PLCC$\uparrow$\\ \midrule
  \multirow{2}{*}{\textbf{KoNViD-1k}}& Temporal  &0.530 &\textbf{0.527}&\textbf{0.531} &\textbf{0.527}&0.524 &0.519 &0.506 &0.503&0.473 &0.475 \\
  ~& Overall &0.692 &\textbf{0.694} &\textbf{0.693} &0.693&0.685 &0.685 &0.677 &0.678&0.668  &0.667 \\\midrule
  \multirow{2}{*}{\textbf{LIVE-VQC}}& Temporal  &0.595 &0.617 &\textbf{0.596} &\textbf{0.618}&\textbf{0.596} &0.609 &0.583 &0.597 &0.544 &0.565 \\
  ~& Overall &0.714 &0.726 &\textbf{0.718} &\textbf{0.730}&0.710 &0.726 &0.699 &0.717&0.686 &0.705  \\ \midrule
  \multirow{2}{*}{\textbf{CVD2014}}&  Temporal &0.400 &0.467&0.408 &0.469&0.450 &0.504&\textbf{0.470} &\textbf{0.534} &0.452 &0.505  \\
  ~&  Overall &0.502 &0.569 &\textbf{0.524} &\textbf{0.597}&0.521 &0.585 &0.511 &0.582&0.492 &0.577  \\\midrule
  \multirow{2}{*}{\textbf{YouTube-UGC}}&  Temporal  &\textbf{0.281} &\textbf{0.326}&0.269 &0.311 &0.277 &0.319&0.261 &0.312 &0.248 &0.311  \\
  ~&  Overall &\textbf{0.241} &\textbf{0.305} &0.228 &0.287&0.208 &0.258 &0.198 &0.235&0.198 &0.235  \\
    \bottomrule
    \end{tabular}
    }
    \vspace{-3mm}
    
 \label{dimension}
\end{table*}

\tabcolsep=0.5pt
\begin{figure*}[t]
	\centering
  \linespread{0.9}
  	\small{ 
		\begin{tabular}{ccc}
		    \includegraphics[width=0.31\textwidth]{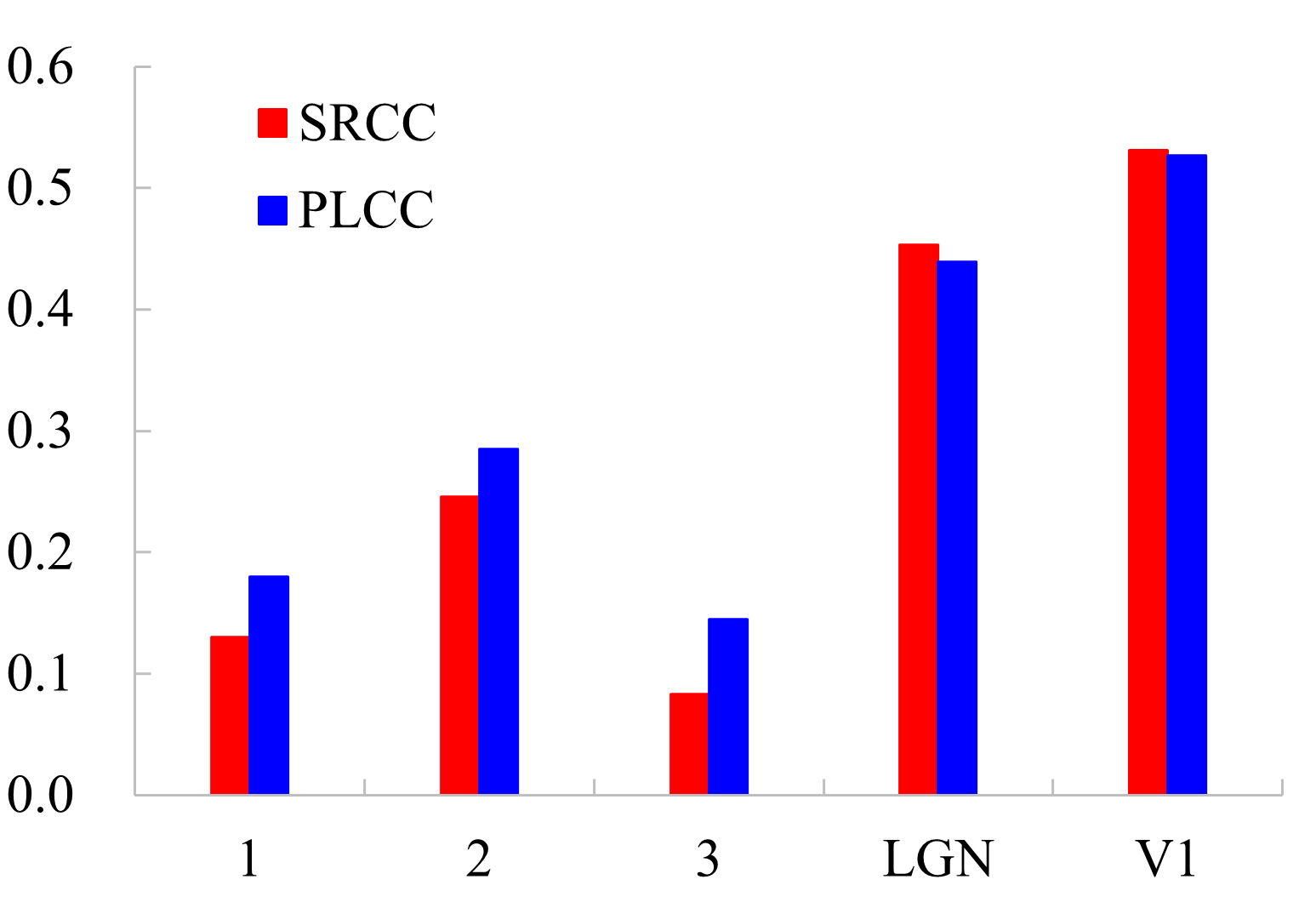}&
		    \includegraphics[width=0.31\textwidth]{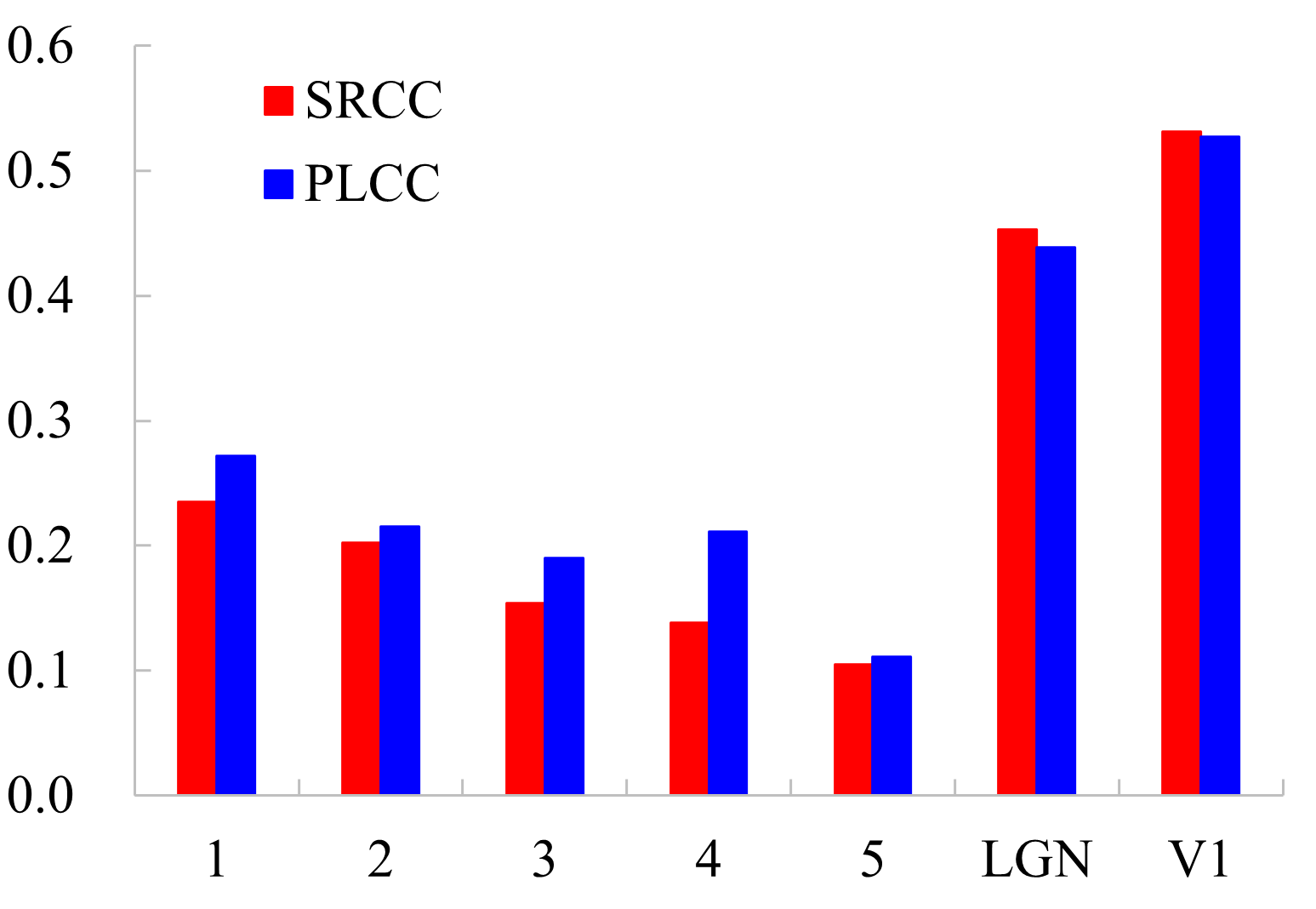} &
		    \includegraphics[width=0.31\textwidth]{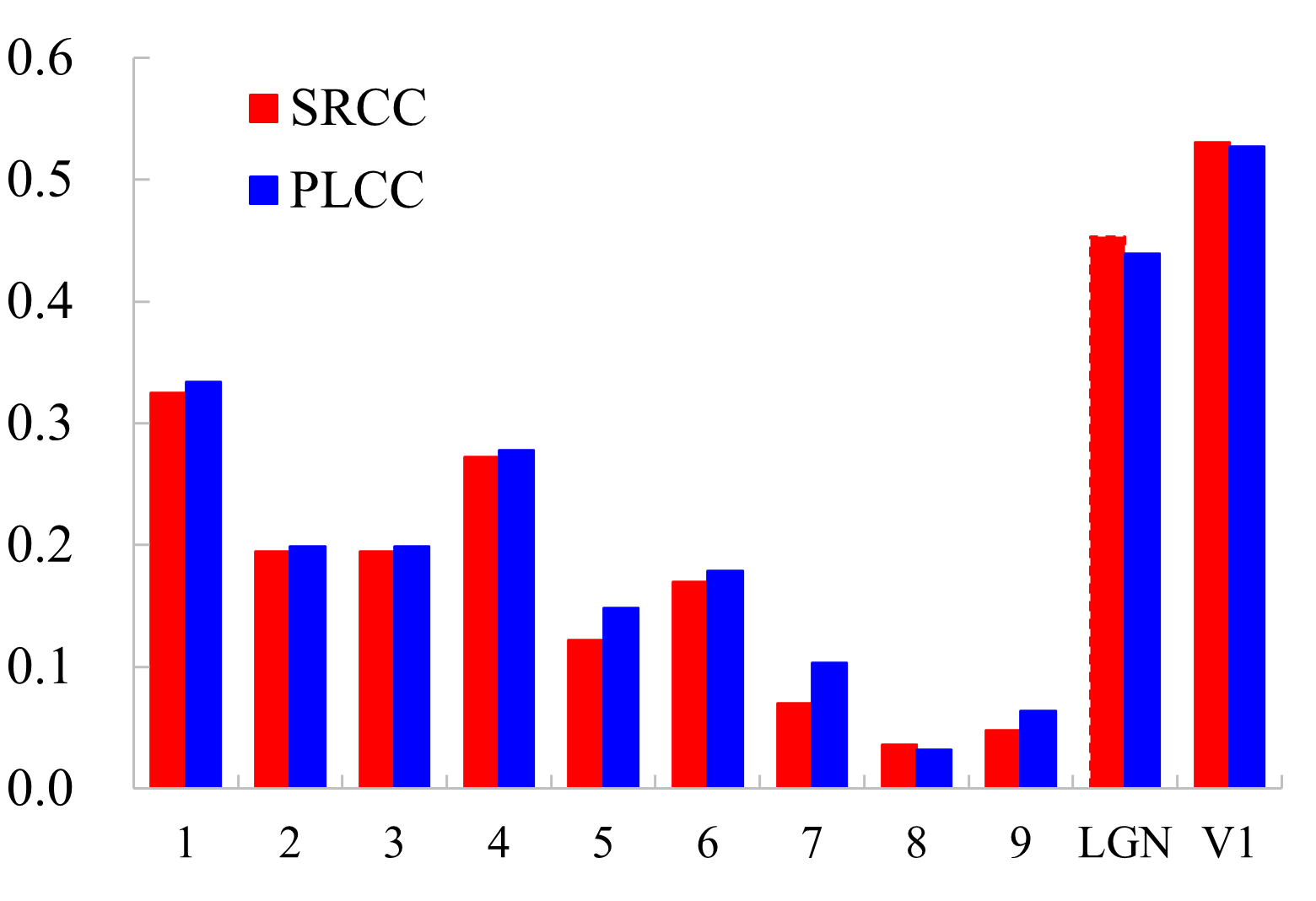} \\
			(a) \emph{1}-th to \emph{3}-th feature blocks of AlexNet& (b) \emph{1}-th to \emph{5}-th feature blocks of VGG19& (c) \emph{1}-th to \emph{9}-th feature blocks of ResNet152 \\
	\end{tabular} }
	\vspace{-4mm}
\caption{\small Performance comparison between bio-inspired features, \emph{i.e.} features extracted by computational models of LGN and V1, and deep features from different layers of three classical CNN models, which have been well-trained on ImageNet~\cite{imagenet} for classification tasks.}
	\vspace{-4mm}
\label{deep}
\end{figure*}

\subsection{Ablation Study}

\subsubsection{Design of the VPT descriptor}
We propose to describe the loss in the straightness and compactness of perceptual domain representation by two components, namely the curvature representing how the representation deviates from the straight line and the distance representing how fast it deviates over a certain time interval. 

\textbf{Effectiveness of curvature and distance components.} To investigate the validity of these two components, we conduct an ablation study on three variants: a) curvature only; b) distance only; c) a combination of curvature and distance (ours). To make the comparison more comprehensive, we also include d) the linear prediction error for temporal modeling in~\cite{STEM22}. The numerical comparisons are shown in Table.~\ref{tab:des}. As stated in~\cite{nature19} that the curvature of the perceptual representation is vital to discriminating between natural and artificial videos, and it is also more effective than the distance for assessing temporal quality. But taking both curvature and distance results in better performance than just utilizing a single component, showing that the distance may show the intensity of temporal distortion, which also accounts for the low subjective score. The proposed descriptor also achieves better performance than the linear model, indicating that the proposed descriptor can better measure the temporal distortions in the perceptual domain.

\textbf{Options for distance components.} To make a better descriptor, we have tested the possible distance measurement options from the perceptual representation. Table.~\ref{tab:distance} shows all the tested options and their performance. In general, the performance of option measuring the compactness of the trajectory by the magnitude of sum of the two difference vectors $S_i$, \emph{i.e.}  $||\bm{\overrightarrow{x_{i-1} x_{i}}}+\bm{\overrightarrow{x_{i} x_{i+1}}}||$, is better than the other options. According to these results, we use it for the distance in the TPQI algorithm.

\subsubsection{The impact of various settings on V1 feature}

The representation in V1 domain has important contribution on the performance of the proposed TPQI, so that we test various settings including the resolutions for feature extraction and feature dimensionality reduction of the presentation in V1 domain. The experiments are conducted on two models: 1) TPQI (V1) model to eliminate effects from the LGN feature and the spatial quality, and 2) Overall model to check its effects on the final proposed VQA algorithm.

\textbf{Resolutions for extracting V1 feature}. We first test various downsampled video resolutions for V1 feature extraction since temporal modeling may not require a high spatial resolution. We conduct the test on the \textbf{KoNViD-1k} dataset with a unified raw resolution of $960\times540$, and apply the conclusion to other datasets. The results are presented in Table.~\ref{resolution} and Fig.~\ref{fig:resolution}, respectively. For TPQI (V1) model, the best performance is reached at downsampling rate of 1/2 (270P); while for the Overall model, the performance also almost reaches saturation at this resolution. Considering that the computational complexity decreases exponentially with the resolution downsampling, we chose to use 270P for the resolution of the input videos to extract the representation in V1 domain.


\textbf{Dimension of V1 feature.} The representation in V1 domain is the feature after dimensionality reduction of the original V1 feature map. We perform an extensive study on the parameter $d$ for the feature dimensionality reduction, and the results are reported in Table~\ref{dimension}. The best results are mostly achieved at $d=10$, and lower or higher dimensions will cause the degradation of the performance. Therefore, we set $d=10$ for the dimension of the representation in V1 domain in the proposed VQA algorithm.

\subsubsection{Bio-inspired handcrafted feature v.s. deep feature}
As being stated that the convolutional neural networks (CNNs) have shown impressive ability in object recognition and been proposed as candidate models for biological vision~\cite{pnas2014,pcbi2014}, we compare the straightening capabilities of their features with the proposed biological LGN and V1 features for VQA. The experiment is conducted by replacing the LGN and V1 features with the features extracted in each stage of the classical CNN models, including Alexnet~\cite{Alexnet}, VGG19~\cite{vgg19}, and Resnet152~\cite{Resnet}. The results in Fig.~\ref{deep} show that the proposed biological features can outperform those deep features extracted from the CNN models in temporal quality perception, which motivate us to address the related vision problems, such as image restoration~\cite{Liao2020ECCV,Liao2021CVPR}, action recognition~\cite{zhong,zhongvideo}, and video compression~\cite{xiaoSPE,xiaoTMM} considering the characteristics of the HVS.



\section{Conclusions}
In summary, we have applied the perceptual straightening hypothesis of the HVS to design a blind temporal quality prediction algorithm called TPQI. We demonstrate the efficacy of TPQI by its superior performance over a number of in-the-wild datasets. The performance of TPQI is noteworthy since it even surpasses supervised VQA algorithms on related datasets. Importantly, temporal consistency checks introduced by this hypothesis play a key role in achieving performance gains in video quality prediction. The proposed TPQI algorithm is explainable and generalizes well over a variety of in-the-wild datasets.

\begin{acks}
This study is supported under the RIE2020 Industry Alignment Fund – Industry Collaboration Projects (IAF-ICP) Funding Initiative, as well as cash and in-kind contribution from the industry partner(s).
\end{acks}


\end{document}